\documentclass[sn-mathphys]{sn-jnl}

\jyear{2021}

\newcommand{\ie}{\emph{i.e.}}

\usepackage{amsmath}
\usepackage{color}
\usepackage{xcolor}

\begin{document}

\title[Article Title]{GRAN: Ghost Residual Attention Network for Single Image Super Resolution}

\author[1]{\fnm{Axi} \sur{Niu}}\email{nax@mail.nwpu.edu.cn}

\author[1]{\fnm{Pei} \sur{Wang}}\email{wangpei23@mail.nwpu.edu.cn}

\author[1]{\fnm{Yu} \sur{Zhu}}\email{yuzhu@nwpu.edu.cn}
\author*[2]{\fnm{Jinqiu} \sur{Sun}}\email{sunjinqiu@nwpu.edu.cn}
\author[1]{\fnm{Qingsen} \sur{Yan}}\email{yqs@mail.nwpu.edu.cn}
\author[1]{\fnm{Yanning} \sur{Zhang}}\email{ynzhang@nwpu.edu.cn}

\affil[1]{\orgdiv{School of Computer Science}, \orgname{Northwestern Polytechnical University}, \orgaddress{\city{Xi’an}, \postcode{710072}, \state{State}, \country{China}}}

\affil[2]{\orgdiv{School of Astronautics}, \orgname{,Northwestern Polytechnical University}, \orgaddress{\city{Xi’an}, \postcode{710072}, \state{State}, \country{China}}}

\abstract{Recently, many works have designed wider and deeper networks to achieve higher image super-resolution performance. Despite their outstanding performance, they still suffer from high computational resources, preventing them from directly applying to embedded devices. To reduce the computation resources and maintain performance, we propose a novel Ghost Residual Attention Network (GRAN) for efficient super-resolution. This paper introduces Ghost Residual Attention Block (GRAB) groups to overcome the drawbacks of the standard convolutional operation,~\ie, redundancy of the intermediate feature. GRAB consists of the Ghost Module and Channel and Spatial Attention Module (CSAM) to alleviate the generation of redundant features. 
Specifically, Ghost Module can reveal information underlying intrinsic features by employing linear operations to replace the standard convolutions. Reducing redundant features by the Ghost Module, our model decreases memory and computing resource requirements in the network. The CSAM pays more comprehensive attention to where and what the feature extraction is, which is critical to recovering the image details. Experiments conducted on the benchmark datasets demonstrate the superior performance of our method in both qualitative and quantitative. Compared to the baseline models, we achieve higher performance with lower computational resources, whose parameters and FLOPs have decreased by more than \textit{ten times}.}


\maketitle

\section{Introduction}

Single Image super-resolution (SISR) is a classic ill-posed inverse problem. SISR aims to obtain a high-resolution (HR) image containing great details and textures from a low-resolution (LR)image by a super-resolution method. It can solve the problems of reduced spatial resolution of imaging results and loss of high-frequency detail information of the scene during the image acquisition process of traditional imaging equipment. Traditional super-resolution research has achieved many wonderful results from simple interpolation to statistical~\cite{2010Super}, self-similar image patch learning~\cite{Yu2014A}, neighborhood embedding~\cite{Xiaoqiang2014Alternatively}, sparse coding~\cite{Yang2010Image}, image patch regression~\cite{2014Anchored}, random forest~\cite{2015Fast} and Bayes~\cite{2016Naive}, etc. However, SISR is still a challenging unsuitable, and pathological task because one specific LR image can correspond to multiple HRs~\cite{niu2023cdpmsr}. Super-resolution is a process of forcibly recovering high-frequency information using low-frequency information.
\begin{figure}[!ht]
	\centering
	\includegraphics[width=7.16cm]{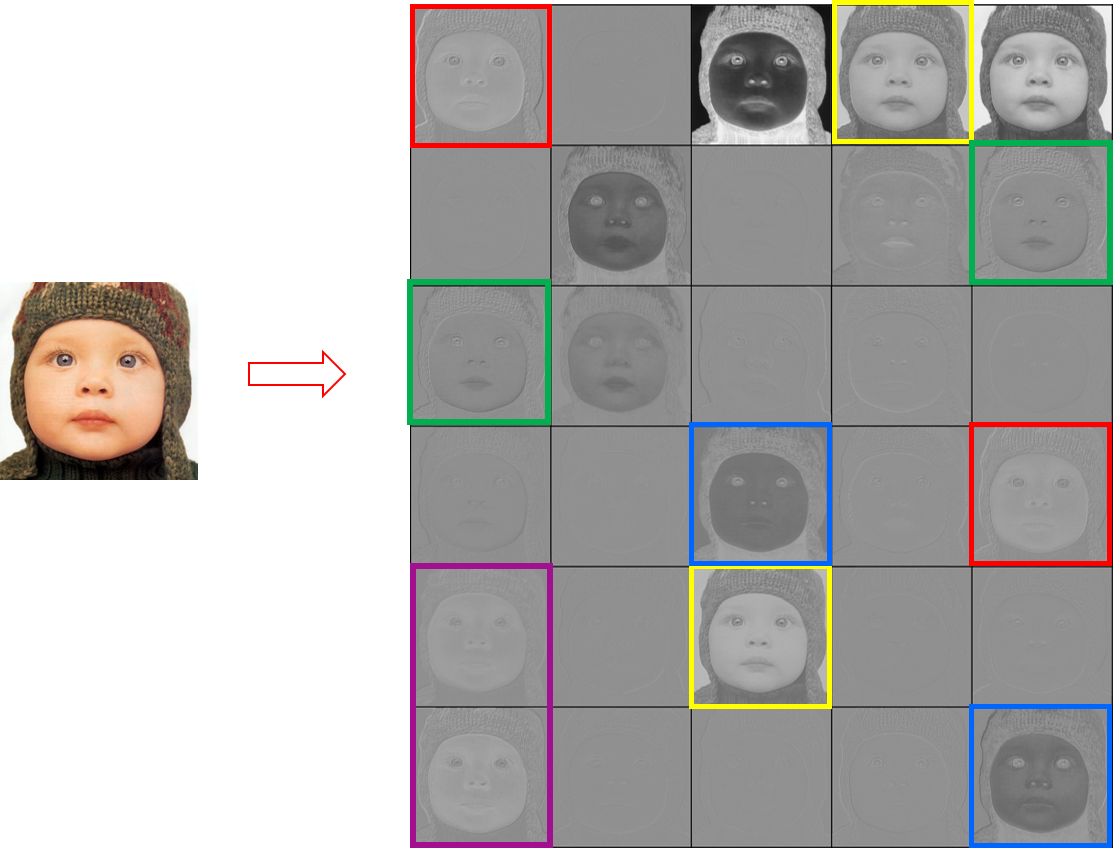}
	\caption{An example of redundant features from the image 'baby' in Set5 obtained by RCAN\cite{zhang2018image}. The are many similar feature pairs in a feature space.}
	\label{feature_RCAN}
\end{figure}

With the wide application of deep learning technology in various fields~\cite{singh2020survey,shamsolmoali2019deep,li2019arsac,li2020enhancing,niu2022fast,mao2023enhancing}, methods based on neural networks have gradually become the mainstream solution to the image super-resolution, which can solve SISR problem by implicitly learning the complex nonlinear-LR-to-HR-mapping-relationship based on numerous LR-HR image pairs~\cite{9044873}. SRCNN~\cite{2016Image} is the first one using deep learning technology to solve the SISR problem. Compared with many traditional SISR methods based on machine learning, the simple structure of the SRCNN model shows remarkable performance in image super-resolution problems. Then, a large number of CNN-based models were proposed to obtain more accurate SISR results using different techniques to improve the quality of the reconstructed image: the design of the network structure with residuals~\cite{barzegar2020super,song2021gradual,peng2020super,hu2020accurate,yang2021wide}; generative adversarial networks~\cite{2018ESRGAN}; neural architecture search~\cite{2019Fast,chu2020multi}; various attention mechanisms \cite{2019Channel}, and other technologies~\cite{ai2020single,liu2020deep}. With the improvement of architecture, this field has indeed made rich progress.

However, these algorithms have higher requirements for hardware resources with their increasingly complex structures, leading them not to be widely used and cannot further improve performance. Considering the assistance provided by the redundant features (see Figure~\ref{feature_RCAN}) for image super-resolution is the same, and with increasing the depth and width of the network, the redundant features will consume a large part of the resources. In this paper, we propose a novel Ghost Residual Attention Network (GRAN),  which can reduce redundant features and decrease hardware requirements while maintaining the network's performance. More specifically, we study (1) designing the lightweight block, Ghost Residual Attention Block (GRAB), and (2) designing better performance feature extraction modules, Channel and Spatial Attention Module (CSAM). 
In summary, our main contributions are as follows: 
\begin{itemize}
    \item We propose an image super-resolution algorithm, which can make the network structure lighter while ensuring the algorithm's effectiveness.
    \item We design a Ghost Residual Attention Block by applying ghost technology, which employs a series of linear operations to replace the traditional convolutions. This operation can significantly reduce redundant features, reducing the network's demand for memory and computing resources.
    \item We redesign a Channel and Spatial Attention Module (CSAM) based on an attention mechanism to comprehensively extract high-frequency features conducive to image reconstruction from space and dimensions.
    \item  Quantitative experiments show that our method not only has a performance advantage, but the number of parameters and calculations also are significantly less than the state of art methods.
\end{itemize}

The rest of the paper is organized as follows: section 2 briefly concludes the related work in the area, followed by the proposed lightweight model GRAN in section 3, the experiments and analysis in section 4, and finally, the conclusion in section 5.

\section{Related Works}\label{sec2}
With the development of deep learning, countless excellent networks for image-processing tasks have been proposed. Along
with it, many SISR technologies combined with deep learning have emerged. Here, we discuss CNN-based methods, lightweight networks, and works related to attention mechanisms.
\subsection{CNN-based methods}
\begin{figure}
	\centering
	\includegraphics[width=7.16 cm]{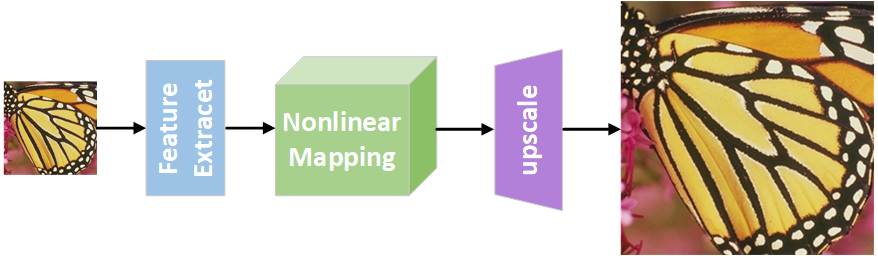}
	\caption{The basic structure of the networks for super-resolution.}
	\label{base_SRnet}
\end{figure}

Since neural networks were introduced into the super-resolution field, the image super-resolution model structure is mostly divided into three parts \cite{2017Enhanced, Li2018Multi,lyn2020multi}, etc., as shown in the Figure~\ref{base_SRnet}: 1) the feature extraction part: to obtain the shallow features of the image by a convolution operation, 2)the nonlinear mapping part: to process the shallow features to obtain the information required for reconstruction with various network structure designs, 3)and the upscale part: to reconstruct the image using a special upscale technology.

Among them, most researchers focused on improving the second part of the network to improve the algorithm's performance. SRCNN~\cite{2016Image} is the first deep-learning work for SISR. There were three convolution layers in SRCNN, corresponding to the three steps: feature extraction, nonlinear mapping, and restoration. After SRCNN, VDSR~\cite{Kim2016Accurate} applied a very deep neural network model combined with residual learning to perform super-scores, which greatly accelerated the model's training process and made the details retained in the final output better. EDSR~\cite{2017Enhanced}removed some unnecessary modules in the residual structure and set up a multi-scale model, which presented a good effect in processing each single-scale super-resolution. RCAN~\cite{zhang2018image} proposed a new residual dense network, RIR(Residual In Residual)technology, using the hierarchical features of all convolutional layers to improve the algorithm's performance. MCAN~\cite{ma2019matrix} introduced a SISR network called Matrix Channel Attention Network (MCAN) by constructing a matrix set of multi-connected channel attention blocks (MCAB), which achieved a fine performance by using fewer multiplications and parameters.~\cite{rajput2020robust} proposes a face image super-resolution reconstruction method based on a combined representation learning method, adopting deep residual and deep neural networks as generators and discriminators. MLRN~\cite{lyn2020multi} proposed a multi-scale fusion architecture to solve the problem that the existing SISR could not make full use of the characteristic information of the network middle layer.~\cite{chen2021face} proposed inheriting the merits of functional-interpolation and dictionary-based SR techniques to produce more discriminate and noise-free high-resolution (HR) images from captured noisy LR probe images is suitable for real-world low-resolution face recognition.~\cite{li2021single} proposes a novel network  for SR called a super-resolution network by fusing internal and external features (SRNIF), which could make full use of the internal features of the image and obtain the detailed texture of the input LR.

The advent of these algorithms has greatly developed the field of super-resolution. However, the redundant feature is an obvious feature of these successful CNNs, but they are rarely considered when designing the network structures. Furthermore, redundant features make the consumption of computing resources meaningless. 

\subsection{Lightweight Network }
The complexity of the network structure and the demanding requirements for computing resources resulted in various network Lightweight technologies: Pruning~\cite{2015Deep,li2016pruning,peng2019collaborative}, Sparse Representation~\cite{guo2018sparse,fan2020neural}, Knowledge distillation~\cite{hinton2015distilling}.~\cite{li2016pruning} analyzed that CNN usually had significant redundancy between different filters and feature channels. The author proposed to reduce the computational cost of CNN by trimming the Filter.~\cite{peng2019collaborative} adopted the channel pruning to the trained model, which reduced the cumulative error and was suitable for various network structures.~\cite{fan2020neural} further targeted removing useless channels for easier acceleration in practice by sparse representation technology. Specifically,~\cite {fukuda2017efficient} and~\cite{heo2019knowledge} utilize larger models to teach smaller ones, named Knowledge distillation, which improves the performance of smaller models. 

The performance of these methods usually depends on the given pre-trained model. The basic operation of these models and their architecture has further improved them.
\subsection{Attention-based Networks}
Attention mechanisms have been widely used in recent computer vision tasks, such as image classification~\cite{hu2018squeeze}, image captioning~\cite{chen2017sca}, image recognition~\cite{zhao2020exploring}, which is an effective method for biased allocation of the valid part of the input information.~\cite{fu2017look} introduced the attention mechanism, making the entire network structure pay attention to the overall information, and the local information.~\cite {woo2018cbam} combined spatial attention mechanism and channel attention mechanism, which had acquired excellent performance in classification and detection. SAN~\cite{zhao2020exploring} applied the self-attention mechanism, which greatly reduced the parameters and calculations and significantly improved the classification accuracy of the ImageNet dataset compared with the classic convolutional network ResNet.

With the popularity of the attention mechanism, more and more attention-based methods are proposed to improve SR performance. Such as,~\cite{zhang2018image} proposed the residual channel attention network (RCAN) by introducing the channel attention mechanism into a modified residual block for image SR. The channel attention mechanism uses global average pooling to extract channel statistics called first-order statistics.

\begin{figure}
	\centering	
	\includegraphics[width=0.95\textwidth]{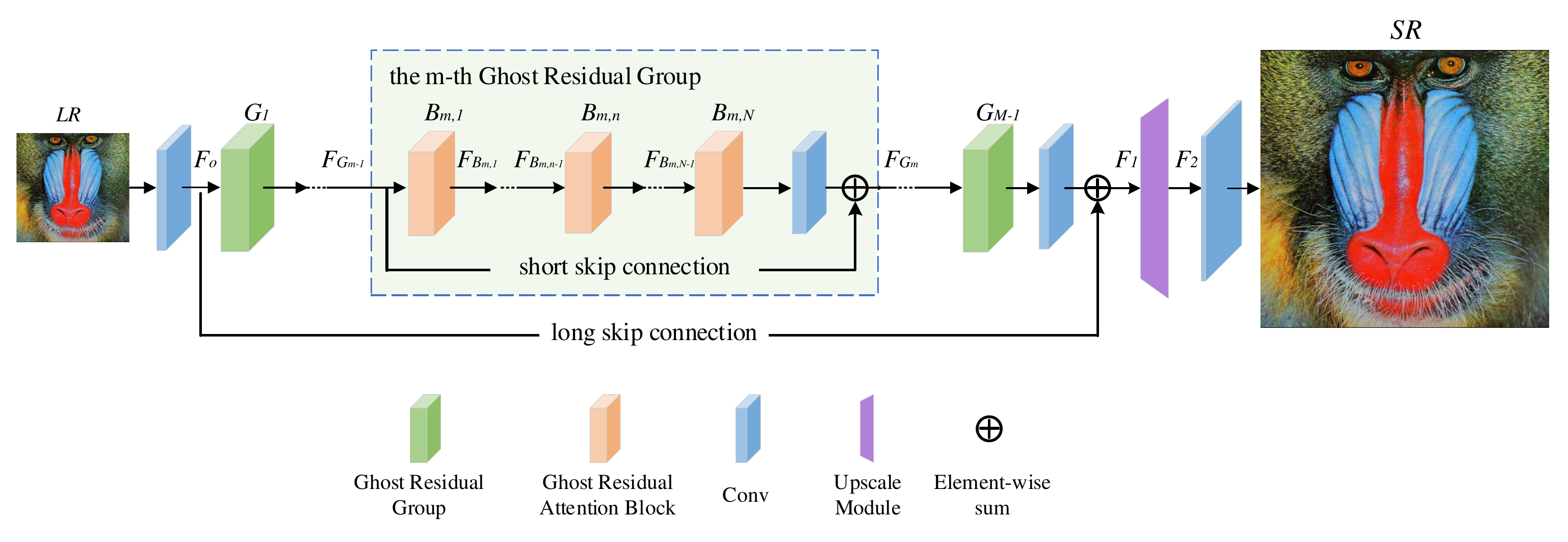}
	\caption{The overview of our GRAN super-resolution network architecture, stacked by multiple GRABs. Each short ship connection divides it into multiple GRABs groups, which can avoid information loss.}
	\label{GRAN}
\end{figure}

\section{Methodology}\label{sec3}
This section focuses on the proposed Ghost Residual Attention Network (GRAN), an end-to-end image super-resolution network. First, we introduce the overall architecture of the network in Section~\ref{sec:network_structure}, and then the main modules of the network in detail in Section~\ref{sec:grab}. Finally, we analyze the complexity of the network in Section~\ref{sec:analysis_complexity}.

\subsection{Network Structure}
\label{sec:network_structure}
The proposed GRAN architecture is shown in Figure~\ref{GRAN}. The GRAN can also be decomposed into three parts: shallow feature extraction module, nonlinear mapping module, and up-scale module. 
At the beginning of the model, we use a single convolution layer as the shallow feature extraction module. Then, our nonlinear mapping module contains multiple Ghost Residual Attention Blocks and has two Ghost Residual Groups in the input and output of this module. Ultimately, our up-scale module combines the output of the shallow feature extraction module and nonlinear mapping module to produce the super-resolution result. 
In formulation, we define the input image and output image of our GRAN as ${I_{LR}}$ and ${I_{SR}}$, respectively. Our GRAN can reconstruct an accurate SR image ${I_{SR}}$ directly from the LR ${I_{LR}}$:
\begin{equation}
    \\I_{SR}=H{_{up}}(H_{map}(H_{extract}(I{_{LR}})))={H_{GRAN}}({I_{LR}}),
\end{equation}
where ${H_{extract}}$, ${H_{map}}$, ${H_{GRAN}}$ represent the feature extraction module, nonlinear mapping module, and up-sample module, respectively. And here, the ${H_{map}}$ represents the module composed of our GRG, which consists of many Ghost Residual Attention Blocks (GRAB)),~\ie the proposed Ghost Residual Attention Block is the backbone of our methods, in the following, we will give details for it how to work.

\subsection{Ghost Residual Attention Block (GRAB)}
\label{sec:grab}
\begin{figure}[ht]
	\centering
	\includegraphics[width=12 cm]{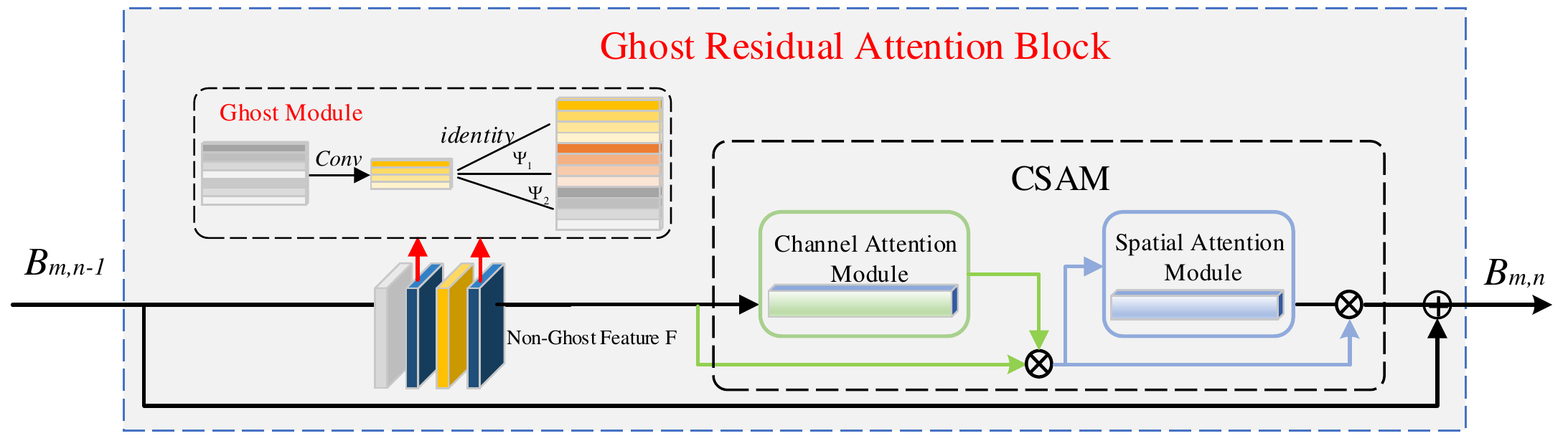}
 \label{fig:GRAB}
	\caption{The overview of GRAB. The left part is the module to remove redundant features, whose convolution operation is replaced by the Ghost module. On the right is the Channel and Spatial Attention Module (CSAM) module, which is obtained by the sequential combination of channel attention and spatial attention.}
	
\end{figure}

Inspired by GhostNet~\cite{2020GhostNet}, we replace the convolution operation in the traditional network structure with a series of linear operations named Ghost Module, as shown in Figure~\ref{fig:GRAB}. The traditional convolutional neural network directly performs convolution operations on the features of each layer. While the Ghost Module proposed by~\cite{2020GhostNet} uses a series of linear operations to operate the feature layer separately, which can introduce more features while reducing the number of network calculations. Here, we use a simplified version of the Ghost model to achieve lightweight and simplification of the network structure. We will conduct a detailed introduction later.

As shown in Figure~\ref{GRAN}, our network is stacked by multiple GRAB groups. The calculation of features in G-th GRAB group via:
\begin{equation}
 \\F_{G} = B_{n}(B_{n-1}(\dots B_{1}(F_{i})\dots )),i=0,,,N
\end{equation}
where, the ${F_0}$ is the shallow feature obtained by ${I_{LR}}$ with a 3×3 convolution, which is used as the input of the first GRAB. $B_{n}$  denotes the n-th GRAB and $F_{i}$(except ${F_0}$) is its output. 

We refer to the RIR structure and divide the GRABs into multiple groups~\cite{zhang2018image}, then connect them through short skip connection (SSC) and long skip connection (LSC), which allows more low-frequency information to be used, and the information flows better. The purpose is to deepen the depth of the network further, and at the same time, the network can better integrate into the training process~\cite{zhang2018image}. The feature of each GRAB Group is calculated via:
\begin{equation}
 \\F_{G} = W_{S}F_{G-1} + F_{B}, 
\end{equation}
where the ${F_{G}}$ denotes the output of the $G$-th GRAB Group. And the ${W_{S}}$ represents the weight set of SSC, which allows the main parts of the network to learn more residual information, and more abundant low-frequency information is easier obtained in the training process. The following equation can express the reconstruction process of the entire network:
\begin{equation}
 \\I_{SR} = H_{up} (F_{0} +W_{L}F_{G}),
   \end{equation}
Similarly, ${W_{L}}$ represents the weight set of LSC. In this paper, for the specific framework, we set the number of Ghost Residual Groups (GRG) as 10. Each GRG consists of 20 Ghost Residual Attention Blocks (GRAB).

\subsubsection{Ghost Module}
\label{sec:ghost_module}

\begin{figure}
	\centering	
	\includegraphics[width=8.5 cm]{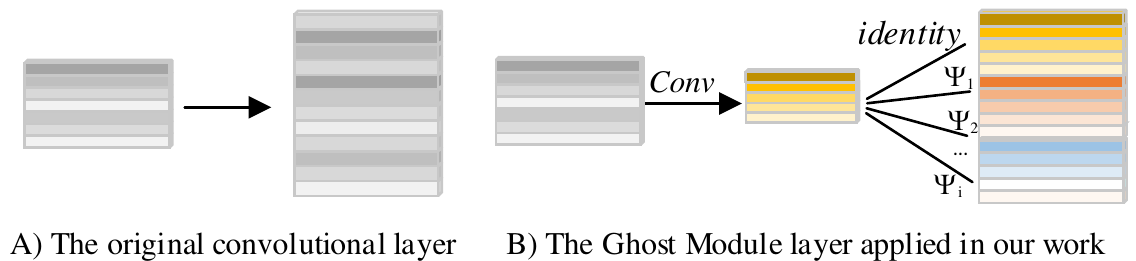}
	\caption{The overview of Ghost Module}
	\label{Ghost_module}
\end{figure}

First, we introduce the Ghost Module (shown in Figure~\ref{Ghost_module}) applied in this article and why it can reduce the complexity of network parameters.
The Ghost technology is first proposed in~\cite{2020GhostNet}. The author proposes to use a variety of linear operations to replace traditional convolution, generate efficient CNNs, and verify the proposed method's effectiveness in ImageNet classification and object detection. Here, we use the Ghost idea to build the Ghost Module and combine it with the attention mechanism to achieve the lightweight and optimization of the super-resolution network structure. 

Generally speaking, the original convolution operation in a traditional network used to generate $n$ feature maps can be formulated as:
\begin{equation}
 \\Y = X \otimes f + b ,
 \label{eq5}
\end{equation}
where $X$ and $Y$ represent the input and output of the convolutional layer, $ \otimes $ represents the convolutional operation, respectively, and n is the kernel number of the convolution. 

While the Ghost Module used in our work can be formulated as follows:
\begin{equation}
\begin{aligned}
\\&{Y}'=X \otimes {f}',\\
   &Y=ID(Y^{'})+\Psi _{1}(Y^{'})+\Psi _{2}(Y^{'}),
  \end{aligned}
   \label{eq6}
\end{equation}

\subsubsection{Channel and Spatial Attention Module (CSAM)}
Inspired by the spatial transformer proposed in STN~\cite{jaderberg2015spatial} and the channel attention in SENet~\cite{hu2018squeeze}, we design a CSAM structure, an attention mechanism module that combines spatial and channel. When given the Non-Ghost Featur  $F$, CSAM sequentially performs the channel attention and spatial attention as illustrated in Figure~\ref{CASM}. The overall attention process can be summarized as follows:

\begin{equation}
\begin{aligned}
\\&F_{ghost}^{'}=M_{c}\odot  F, \\
  &F_{ghost}^{"}=M_{s}\odot  F^{'},
  \end{aligned}
\end{equation}
where $M_{c}$ and $M_{s}$ denote the 1D channel attention map and 2D spatial attention map separately. Here, the $\odot$ indicts element-wise multiplication. $F_{ghost}^{"}$ is the final refined output. Figure ~\ref{CASM} depicts the computation process of each attention map. \cite{2018CBAM} proposed a junction structure that combines spatial and channel attention and applied it to image classification and target detection. We reconsider the regional characteristics of the spatial attention mechanism and the global characteristics of the channel attention mechanism and propose to apply this structure to image super-resolution. The following describes the details of each attention module:

\begin{figure}
	\centering	
	\includegraphics[width=8.5cm]{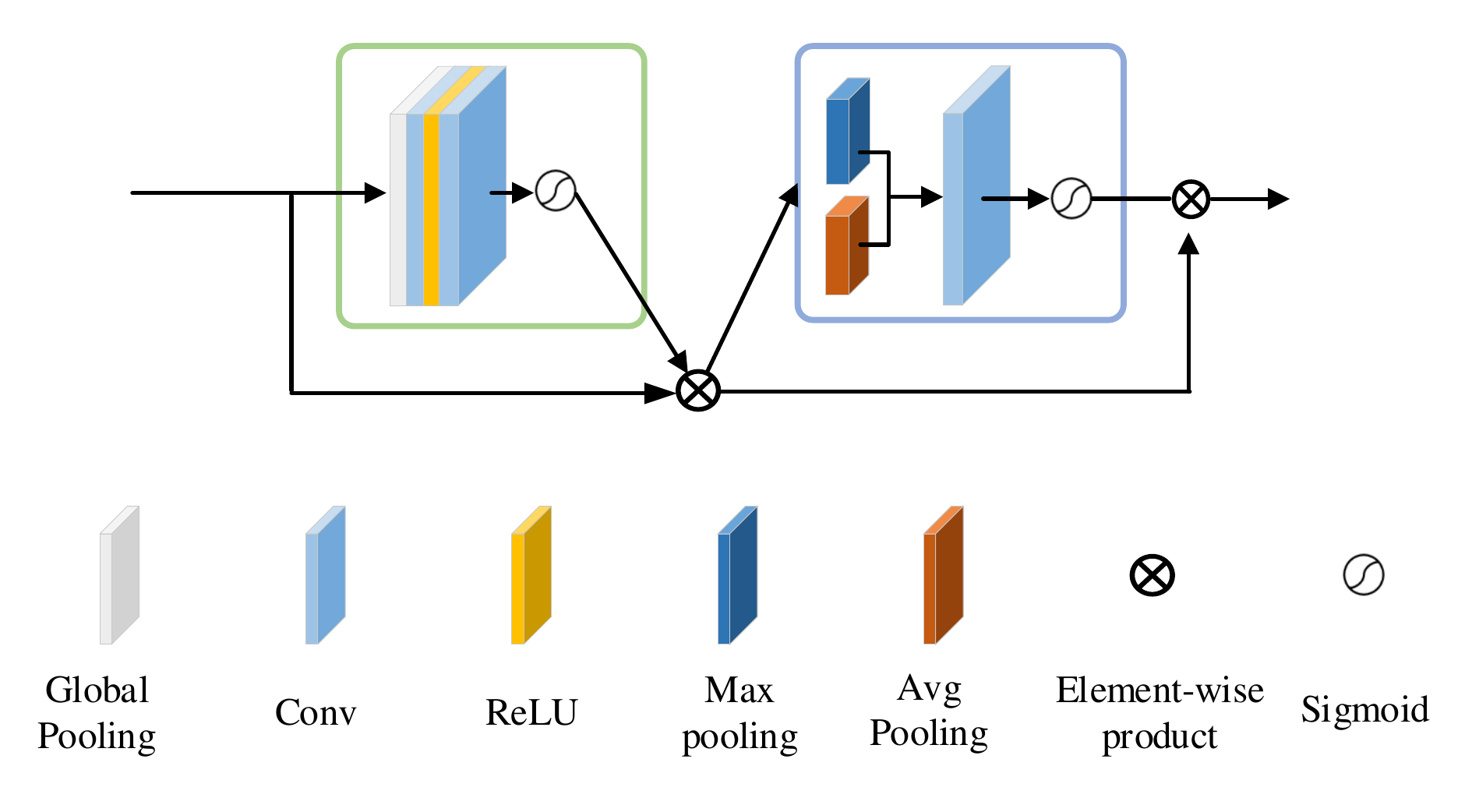}
	\caption{The overview of our CSAM }
	\label{CASM}
\end{figure}

\textbf{Channel attention module.}
This module focuses on selecting useful features and assigning weight to each feature map. Thus this module can help the model know which to look~\cite{hu2018squeeze}. Each layer of the current convolutional neural network has many convolution kernels, corresponding to a feature map. In the channel dimension, different weights are learned for each feature to play its role. Compared with the spatial attention mechanism, channel attention allocates resources between each channel.

\textbf{Spatial attention module.}
This module can be understood as guiding the neural network where to look. Through the attention mechanism, the network can transform the spatial information in the original image into another space and retain the key information. This kind of network is used in many existing methods~\cite{jaderberg2015spatial,2020An}. The network trained through the spatial attention mechanism can find out the area of the image that needs to be paid more attention. At the same time, this mechanism can have the functions of rotation, scaling, and transformation, so that the important information of the part of the picture can be extracted by the box through the transformation.

In addition, our CSAM structure is composed of channel attention and spatial attention in order. This combination order has been proved in~\cite{2018CBAM}, that is, the channel-first order is slightly better than the spatial-first order.

\subsection{Analysis on Complexity}
\label{sec:analysis_complexity}
Compared with RCAN our GRAN achieves higher performance. Meanwhile, our network is more efficient. In this section, we will analyze our network structure from both flops and parameters. Since the computational volume consumption mainly comes from Ghost Residual Attention Block (GRAB) in our method and Residual Channel Attention Block (RCAB) in RCAN, we compare the computation consumption of these two blocks in detail.

\textbf{FLOPs calculation.} 
The main difference between our GRAB and RCAB is that we replace the CNN layer with the Ghost module. Second, following the channel attention module we introduce a spatial attention module which brings negligible computation increase. Here we mainly compute the computation consumption caused by CNN and the channel layer.

For Ghost CNN, the computation process is as follows. First, perform a traditional convolution operation where the convolution kernel is ${f}' \in  R^{c\times 1\times 1\times m } $ on the original feature map $X$ to obtain $m$ intrinsic features ${Y}'$ (Equation~\ref{eq6}), $m\leq n$. Next, generate $q$ features for each feature ${y_{i}}'$ in ${Y}'$  by a series of linear operations:
\begin{equation}
    y_{i,j}=\Psi _{i}({y_{i}}'),   {\rm{ }}\forall i = 1,...,m,{\rm{   }}j = 1,...,q,
\end{equation}
in this way, each ${y_{i}}'\in {Y_{i}}'$ can get one or more feature maps $\{ {y_{i,j}}\} _{j = 1}^q$ by $\Psi _{i}$, the linear operation for generating the $j$-th ghost feature map ${y_{i,j}}$. 

Finally, we can obtain ${n = m \cdot q}$ feature maps ${Y = \left[ {{y_{11}},{\rm{ }}{y_{12}},{\rm{ }} \cdots {\rm{, }}{y_{ms}}} \right]}$ as the output data of Ghost Module. In actual processing, we use identity as the first $\Psi _{1}$ to perform identity mapping on the original feature map. So, we have $m\cdot (q-1)= \frac{n}{q} \cdot (q-1)$ linear operations. Thus, we can get the FLOPs of Ghost CNN as $m\cdot {h}'\cdot {w}'\cdot c\cdot k\cdot k+(q-1)\cdot m\cdot {h}'\cdot {w}'\cdot d\cdot d$.
The FLOPs of a CNN can be simply calculated as $n\cdot {h}'\cdot {w}'\cdot c\cdot k\cdot k$. Considering channel attention spatial attention module brings much lower computation consummation. We simply ignore them. Then we can get the speed ratio $r_q$ as:

\begin{equation}
\begin{array}{rl}
r_{q}=&\frac{n\cdot {h}'\cdot {w}'\cdot c\cdot k\cdot k}{m\cdot {h}'\cdot {w}'\cdot c\cdot k\cdot k+(q-1)\cdot m\cdot {h}'\cdot {w}'\cdot d\cdot d}\\
=&\frac{q\cdot c\cdot k\cdot k}{ c\cdot k\cdot k+({q-1})\cdot d\cdot d}
\end{array}
\end{equation}

Since we set $d\approx k$, and $q\gg c$, we can have:
\begin{equation}
    \\r_{q}\approx\frac{q\cdot c}{c+(q-1)}\approx{q}.
\end{equation}
Here we show the main reason for speedup.
In addition, because CSAM is a lightweight module that can be seamlessly integrated into any CNN architecture, it has almost no impact on efficiency and computing power.

\textbf{Number of parameters.} We mainly analyze the number of parameters used in our Ghost Module and the ordinary convolution layers and ignore the effect of channel attention and spatial attention module because of their simplicity. As introduced in the previous section, the formulation for the original convolution operation can be described as Equation ~\ref{eq5}.

Assuming that the size of the convolution kernel is  $k \times k$, the input channel is $M$, the output channel is $N$, and the number of parameters required for a traditional convolution is :
\begin{equation}
 \\Sum_{conv} =k\times k\times M\times N.
\end{equation}

Here, we ignore the number of biases, which is the same as the number of output channels).

The ghost module used to replace the convolution operation is expressed as:
\begin{equation}
 \\X^{'} =priCONV(X) ,
\end{equation}
\begin{equation}
 \\Y=Identity(X^{'})+ \Psi _{1}(X^{'})+\Psi _{2}(X^{'}) ,
\end{equation}
where $ priCONV $ represent the traditional $1 \times 1$ convolution. Denote $k_{1} \times k_{1}$ and $k_{2} \times k_{2}$ as the kernel size of $\Psi _{1}$ and $\Psi _{2}$ respectively. For the same $M$ and $N$, the parameters of the ghost module should be:
\begin{equation}
\begin{aligned}
 \\Sum_{ghost} &=(1\times 1\times M\times \frac{N}{3}) + (k_{1} \times k_{1} \times \frac{N}{3}\times \frac{N}{3}) \\
 &+(k_{2}\times k_{2} \times \frac{N}{3}\times \frac{N}{3})\\
 &=\frac{N}{3}(M+k_{1} \times k_{1} \times \frac{N}{3}+k_{2} \times k_{2} \times \frac{N}{3})\\
 &\ll Sum_{conv},    \; \; \; \; if( k > k_{1} , k > k_{2} )
 \end{aligned} 
\end{equation}

In most network structures, the values of M and N are relatively large, such as 128, 256, 512, or even 1024. If  $k$ is larger than both $k_{1}$ and $k_{2}$(Set to 1 and 3 respectively in our experiment), $Sum_{ghost}$ will be much smaller than $Sum_{conv}$.

\section{Experiment Results}
\subsection{Implement details}

~~~~\textbf{Datasets and Evaluation Metrics.} Our GRAN is designed to super-resolve LR images, so we use the DIV2K~\cite{timofte2017ntire} as the model's training set, which contains 800 2K high-resolution images for training and 100 images for both validation and testing. For testing, we make comparisons across four scaling tasks (×2, ×3,  ×4, ×8) on datasets: Set5~\cite{bevilacqua2012low}, Set14~\cite{zeyde2010single}, B100~\cite{martin2001database}, Urban100~\cite{huang2015single}, Manga109~\cite{matsui2017sketch}. The evaluation metrics we used are PSNR and SSIM on the Y channel in the YCbCr color space. In addition, we also experiment on real-world LR images (from RealSR~\cite{cai2019toward} dataset) and compare the results qualitatively. Ultimately, we perform some ablation studies to investigate the influence of every component of our GRAB.

\textbf{Training Settings.}
Before training the network, we first perform data augmentation on the 800 training images, by $90^{\circ}$, ${180^\circ }$, ${270^\circ }$ rotation, and horizontal flip. For training, we set the batch size to 12 with the size of 48×48 extracted as inputs, and we apply Adam ($\beta1 = 0.9$, $\beta2 = 0.999$, and = 10-8) to train our model. The initial learning rate is set to $1 \times {10^{-4}}$, halved every $2 \times {10^{-5}}$ steps of back-propagation. And we perform three linear functions to take place of the original, namely identity,${\Psi_{1}}$ (d=3),${\Psi_{2}}(d=5)$. In addition, We use the PyTorch framework to implement our models with a GeForce RTX 1080Ti GPU.

\subsection{Comparisons with State-of-the-art Algorithms}
We compare our method with 11 state-of-the-art SISR methods: SRCNN~ \cite{2016Image}, VDSR~\cite{Kim2016Accurate}, CARN~\cite{2018Fast}, MSRN~\cite{Li2018Multi}, RCAN~\cite{zhang2018image}, SISR-CA-OA \cite{chen2019deep}, MLRN~\cite{lyn2020multi}, MCAN~\cite{ma2019matrix}, DBPN~\cite{haris2019deep}, HRAN~ \cite{muqeet2019hybrid}, ARCAN~\cite{cao2020adaptive}.  These methods are some representative and outstanding methods since 2016, especially since their network structure is becoming more and more complex for improving performance.

{\bfseries Quantitative comparison on PSNR/SSIM.} The quantitative results of different methods on benchmark datasets are provided in Table 1, from which we can find our proposed method has better performance in the current state of the art. We did a lot of experiments on the general datasets for $\times$2, $\times$3, $\times$4, and $\times$8 SR, and use three colors of red, green, and blue to mark the first, second, and third place experimental results. To be fair, all tests were done on a 1080Ti GPU. Some methods have no public code, and the test results are obtained from their original article.

\begin{table}
	\centering
	\caption{\centering Quantitative comparison with the state-of-the-art methods based on ×2, ×3, ×4 SR with bicubic degradation model. The best three results are highlighted in red, green, and blue colors, respectively}
	\label{tab:results_psnr}
	\begin{center}
	\resizebox{\textwidth}{!}{
	\begin{tabular}{c|c|c|c|c|c|c|c|c|c|c|c}
	\toprule
		\hline
		{\multirow{2}{*}{Methods}} &
		\multicolumn{1}{|c|}{\multirow{2}{*}{Scale}} & \multicolumn{2}{c|}{Set5} & \multicolumn{2}{c|}{Set14} & \multicolumn{2}{c|}{B100} & \multicolumn{2}{c|}{Urban100} & \multicolumn{2}{c}{Manga109} \\
		\cline{3-12}
		&                                & PSNR        & SSIM        & PSNR         & SSIM        & PSNR        & SSIM        & PSNR          & SSIM          & PSNR         & SSIM           \\ \hline
		Bicubic                 & $\times 2$                     & 33.66       & 0.9299      & 30.24        & 0.8688      & 29.56       & 0.8431      & 26.88         & 0.8403        & 30.80        & 0.9339         \\
		SRCNN                   & $\times 2$                     & 36.66       & 0.9542      & 32.45        & 0.9067      & 31.36       & 0.8879      & 29.50         & 0.8946        & 35.60        & 0.9663         \\
		VDSR                    & $\times 2$                     & 37.53       & 0.9590      & 33.05        & 0.9130      & 31.90       & 0.8960      & 30.77         & 0.9140        & 37.22        & 0.9750         \\
		CARN                    & $\times 2$                     & 37.72       & 0.9590      & 33.52        & 0.9166      & 32.09       & 0.8978      & 31.92         & 0.9256        & 38.36        & 0.9761         \\
		MSRN                    & $\times 2$                     & 38.07       & 0.9605      & 33.74        & 0.9170      & 32.22       & 0.9013      & 32.22         & 0.9326        & 38.82        & 0.9868         \\
		MLRN                    & $\times 2$                     & 37.90       & 0.9601      & 33.49        & 0.9140      & 32.11       & 0.8989      & 31.87         & 0.9260        &              &                \\
		SISR-CA-OA              & $\times 2$                     & 37.97       & 0.9605      & 33.42        & 0.9158      & 32.15       & 0.8993      & 31.57         & 0.9226        & 38.38        & /0.9755        \\
		MCAN                    & $\times 2$                     & 37.91       & 0.9597      & 33.69        & 0.9183      & 32.18       & 0.8994      & 32.46         & 0.9303        &              &                \\
		DBPN                    & $\times 2$                     & 38.08       & 0.9601      & \textcolor{green}{\textbf{33.85}}       & 0.9190      & 32.27       & 0.9001      & 32.92         & 0.9310        & 39.28        & 0.9770         \\
		HRAN                    & $\times 2$       & \textcolor{green}{\textbf{38.21}}       & \textcolor{blue}{\textbf{0.9613}}      & \textcolor{green}{\textbf{33.85}}        & \textcolor{blue}{\textbf{0.9200}}      & \textcolor{blue}{\textbf{32.33}}      &\textcolor{blue}{\textbf{0.9016}}       & \textcolor{green}{\textbf{32.65}}         & \textcolor{blue}{\textbf{0.9357}}        &\textcolor{blue}{\textbf{39.11}}       &\textcolor{blue}{ \textbf{0.9780}}         \\
		ARCAN                   & $\times 2$                     & 38.01       & 0.9605      & 33.54        & 0.9173      & 32.15       & 0.8992      & 32.13         & 0.9276        & 38.70        & 0.9750         \\
		RCAN                   & $\times 2$       &\textcolor{red}{\textbf{38.27}}        & \textcolor{green}{\textbf{0.9614}}      & \textcolor{red}{\textbf{34.12}}        &\textcolor{green}{\textbf{0.9216}}     &\textcolor{red}{\textbf{32.41}}       &\textcolor{green}{\textbf{0.9027}}       &\textcolor{red}{\textbf{33.34}}          & \textcolor{green}{\textbf{0.9384}}      &\textcolor{red}{\textbf{39.44}}         &\textcolor{green}{\textbf{0.9786}}          \\
		GRAN(ours)             & $\times 2$      & \textcolor{blue}{\textbf{38.16}}      & \textcolor{red}{\textbf{0.9654}}     &\textcolor{green}{\textbf{33.85}}       & \textcolor{red}{\textbf{0.9233}}     & \textcolor{green}{\textbf{32.35}}       &\textcolor{red}{\textbf{0.9042}}   & \textcolor{blue}{\textbf{32.46}}        & \textcolor{red}{\textbf{0.9388}}        & \textcolor{green}{\textbf{39.16}}    & \textcolor{red}{\textbf{0.9815}}         \\ \hline
		Bicubic                 & $\times 3$                     & 30.39       & 0.8682      & 27.55        & 0.7742      & 27.21       & 0.7385      & 24.46         & 0.7349        & 26.95        & 0.8556         \\
		SRCNN                   & $\times 3$                      & 32.75       & 0.9090      & 29.30        & 0.8215      & 28.41       & 0.7863      & 26.24         & 0.7989        & 30.48        & 0.9117         \\
		VDSR                    & $\times 3$                      & 33.67       & 0.9210      & 29.78        & 0.8320      & 28.83       & 0.7990      & 27.14         & 0.8290        & 32.01        & 0.9340         \\
		CARN                    & $\times 3$                      & 34.29       & 0.9542      & 0.9542       & 0.8407      & 29.06       & 0.8034      & 28.06         & 0.8493        & 33.50        & 0.9360         \\
		MSRN                    & $\times 3$                      & 34.38       & 0.9262      & 30.34        & 0.8395      & 29.08       & 0.8041      & 28.08         & 0.8554        & 33.44        & 0.9427         \\
		MLRN                    & $\times 3$                      & 34.18       & 0.9254      & 30.22        & 0.8369      & 29.01       & 0.8033      & 27.88         & 0.8469        &              &                \\
		SISR-CA-OA              & $\times 3$                      & 34.23       & 0.9261      & 30.05        & 0.8363      & 29.01       & 0.8023      & 27.67         & 0.8403        & 32.92        & 0.9391         \\
		MCAN                    & $\times 3$                      & 34.45       & 0.9271      & 30.43        & 0.8433      & 29.14       & 0.8060      & 28.47         & 0.8580        &              &                \\
		HRAN                    & $\times 3$                      &\textcolor{blue}{\textbf{34.69}}       & 0.9292      & \textcolor{green}{\textbf{30.54}}        &\textcolor{green}{\textbf{0.8463}}       & \textcolor{blue}{\textbf{29.25}}       & \textcolor{green}{\textbf{0.8089}}      & \textcolor{green}{\textbf{28.76}}         & \textcolor{green}{\textbf{0.8645}}        & \textcolor{green}{\textbf{34.08}}        & \textcolor{green}{\textbf{0.9479}}         \\
		ARCAN                   & $\times 3$                      & 34.36       & \textcolor{red}{\textbf{0.9542}}     & 30.30        & 0.8412      & 29.07       & 0.8045      & 28.14         & 0.8514        & 33.50        & 0.9439         \\
		RCAN                 & $\times 3$         
		&\textcolor{red}{\textbf{34.74}}      &\textcolor{blue}{\textbf{0.9299}}   &\textcolor{red}{\textbf{30.65}}      &\textcolor{red}{\textbf{0.8482}}   &\textcolor{red}{\textbf{29.32}}      &\textcolor{red}{\textbf{0.8111}}   &\textcolor{red}{\textbf{29.09}}      &\textcolor{red}{\textbf{0.8702}}   &\textcolor{red}{\textbf{34.44}}      &\textcolor{red}{\textbf{0.9499}}          \\
		GRAN(ours)           & $\times 3$        & \textcolor{green}{\textbf{34.58}}      & \textcolor{green}{\textbf{0.9302}}      & \textcolor{blue}{\textbf{30.53}}       & \textcolor{blue}{\textbf{0.8455}}    & \textcolor{green}{\textbf{29.29}}       & \textcolor{blue}{\textbf{0.8079}}      & \textcolor{blue}{\textbf{28.31}}     & \textcolor{blue}{\textbf{0.8623}}     & \textcolor{blue}{\textbf{33.96}}   & \textcolor{blue}{\textbf{0.9476}}    \\ \hline
		Bicubic                 & $\times 4$                      & 28.42       & 0.8104      & 26.00        & 0.7027      & 25.96       & 0.6675      & 23.14         & 0.6577        & 24.89        & 0.7866         \\
		SRCNN                   & $\times 4$                     & 30.48       & 0.8628      & 27.50        & 0.7513      & 26.90       & 0.7101      & 24.52         & 0.7221        & 27.58        & 0.8555         \\
		VDSR                    & $\times 4$                     & 31.35       & 0.8830      & 28.02        & 0.7680      & 27.29       & 0.7251      & 25.18         & 0.7540        & 28.83        & 0.8870         \\
		CARN                    & $\times 4$                     & 32.13       & 0.8937      & 28.60        & 0.7806      & 27.58       & 0.7349      & 26.07         & 0.7837        & 30.47        & 0.9084         \\
		MSRN                    & $\times 4$                     & 32.07       & 0.8903      & 28.60        & 0.775       & 27.52       & 0.7273      & 26.04         & 0.7896        & 30.17        & 0.9034         \\
		MLRN                    & $\times 4$                     & 31.92       & 0.8911      & 28.43        & 0.7748      & 27.49       & 0.7334      & 25.78         & 0.7763        &              &                \\
		SISR-CA-OA              & $\times 4$                     & 31.88       & 0.8900      & 28.31        & 0.7740      & 27.45       & 0.7303      & 25.56         & 0.7670        & 29.61        & 0.8944         \\
		MCAN                    & $\times 4$                     & 32.33       & 0.8959      & 28.72        & 0.7835      & 27.63       & 0.7378      & 26.43         & 0.7953        &              &                \\
		DBPN                    & $\times 4$                     &\textcolor{red}{\textbf{32.65}}        & \textcolor{blue}{\textbf{0.8990}}       & \textcolor{red}{\textbf{29.03}}        &\textcolor{red}{\textbf{0.7910}}       & \textcolor{green}{\textbf{27.82}}                & \textcolor{red}{\textbf{0.7440}}   & \textcolor{red}{\textbf{27.08}}   &\textcolor{red}{\textbf{0.8140}}         & \textcolor{red}{\textbf{31.74}}   & \textcolor{red}{\textbf{0.921}}         \\
		HRAN                    & $\times 4$                     & 32.43       & 0.8976      & 28.76        & \textcolor{blue}{\textbf{0.7863}}      & 27.70       & \textcolor{blue}{\textbf{0.7407}}  & \textcolor{blue}{\textbf{26.55}}   & \textcolor{blue}{\textbf{0.8006}}  & \textcolor{blue}{\textbf{30.94}}   & 0.9143         \\
		ARCAN                   & $\times 4$                     & 32.13       & 0.8941      & 28.58        & 0.7816      & 27.56       & 0.7356      & 26.09         & 0.7859        & 30.42        & 0.9074         \\
		RCAN                   & $\times 4$                     &\textcolor{green}{\textbf{32.63 }}       & \textcolor{green}{\textbf{0.9002}}      &\textcolor{blue}{\textbf{28.87}}        & \textcolor{green}{\textbf{0.7889}}      & \textcolor{blue}{\textbf{27.77}}       &\textcolor{green}{\textbf{0.7436}}       &\textcolor{green}{\textbf{26.82}}          &\textcolor{green}{\textbf{0.8087}}         & \textcolor{green}{\textbf{31.22}}        & \textcolor{green}{\textbf{0.9173}}         \\
		GRAN(ours)                   & $\times 4$                     & \textcolor{blue}{\textbf{32.54}}       & \textcolor{red}{\textbf{0.9114}}      &  \textcolor{green}{\textbf{28.88}}    &\textbf{0.7858}      & \textcolor{red}{\textbf{27.84}}       & \textbf{0.7276}   &\textbf{26.51}       &\textbf{0.7995}       & \textbf{30.59}   &\textcolor{blue}{\textbf{0.9152}}        \\  \hline
		Bicubic                 & $\times 8$                     & 24.40       & 0.6580      & 23.10        & 0.5660      & 23.67       & 0.5480      & 20.74         & 0.5160        & 21.47        & 0.6500         \\
		SRCNN                   & $\times 8$                     & 25.33       & 0.6900      & 23.76        & 0.5910      & 24.13       & 0.5660      & 21.29         & 0.5440        & 22.46        & 0.6950         \\
		VDSR                    & $\times 8$                     & 25.93       & 0.7240      & 24.26        & 0.6140      & 24.49       & 0.5830      & 21.70         & 0.5710        & 23.16        & 0.7250         \\
		MSRN                    & $\times 8$                     & 26.59       & 0.7254      & 24.88        & 0.5961      & 24.70       & 0.5410      & 22.37         & 0.5977        & 24.28        & 0.7517         \\
		DBPN           & $\times 8$       & \textcolor{red}{\textbf{27.51}}       &\textcolor{red}{\textbf{0.793}}       & \textcolor{red}{\textbf{25.41}}       & \textcolor{red}{\textbf{0.657}}       & \textcolor{red}{\textbf{25.05}}       &\textcolor{red}{\textbf{0.607}}       & \textcolor{red}{\textbf{23.20}}         &\textcolor{red}{\textbf{0.652}}          & \textcolor{red}{\textbf{25.71}}       &\textcolor{red}{\textbf{0.813}}          \\
		HRAN          & $\times 8$   & \textcolor{blue}{\textbf{27.11}}       & \textcolor{blue}{\textbf{0.7798}}      & \textcolor{blue}{\textbf{25.01}}        & \textcolor{blue}{\textbf{0.6419}}      & \textcolor{blue}{\textbf{24.83}}       & \textcolor{blue}{\textbf{0.5983}}      & \textcolor{blue}{\textbf{22.57}}         & \textcolor{blue}{\textbf{0.6223}}        & \textcolor{blue}{\textbf{24.64}}        &\textcolor{blue}{\textbf{0.7817}}         \\
		RCAN                   & $\times 8$    &\textcolor{green}{\textbf{27.31}}        & \textcolor{green}{\textbf{0.7878}}      & \textcolor{green}{\textbf{25.23}}    &\textcolor{green}{\textbf{0.6511}}   & \textcolor{green}{\textbf{24.98}} &\textcolor{green}{\textbf{0.6058}}       &\textcolor{green}{\textbf{23.00}}          & \textcolor{green}{\textbf{0.6452}}        &\textcolor{green}{\textbf{25.24}}       & \textcolor{green}{\textbf{0.8029}}         \\
		GRAN(ours)                    & $\times 8$                     & \textbf{27.01}       &\textbf{0.7735}       &\textbf{24.98}         & \textbf{0.6401}      & \textcolor{blue}{\textbf{24.83}}       &\textbf{0.5978}       & \textbf{22.55}         & \textbf{0.6209}        &\textbf{24.51}         &\textbf{0.7758}          \\ 
		\hline
		\bottomrule
	\end{tabular}}
	\end{center}
	
\end{table}
 
As shown in Table~\ref{tab:results_psnr}. The Bicubic method just interpolates the image directly, so the result is not good. SRCNN is the first method to apply neural networks for super-resolution. VDSR, MSRN, MLRN, SISR-CA-OA, MCAN, DBPN, HRAN, ARCAN, and RCAN, are all excellent super-resolution algorithms from 2017 so far. They have made different improvements by designing the deeper structure, new network unit, feature fusion mode, or connection between network layers, but they all have ignored the redundant features' impact. Therefore, even if the network performance is slightly improved, the parameters and the computation are still shortcomings. In addition, the method HRAN obtains a super-resolution network with excellent performance by searching the network structure, which has stricter consumption of computing resources and time. 

Based on the consideration of redundant features, we propose our lightweight model GRAN, which uses Ghost technology to replace the traditional convolution to reduce the generation of redundant features, thereby reducing the number of parameters and calculations of the algorithm model. In addition, We also add the currently popular attention mechanism (CSAM) to our model to achieve the purpose of super-resolution by fully utilizing the spatial and dimensional information of the image.

In the case of various data sets and different reconstruction sizes, though the value of individual indicators obtained by our method is slightly lower than the optimal algorithm RCAN, HRAN, our parameter quantity has obvious advantages. See from Table~\ref{tab:flop_param}. We have done a lot of comparisons on the benchmark data set for the indicators PSNR and SSIM on the x2, x3, x4, and x8 scales. Red, green, and blue represent the first, second, and third places respectively. Obviously, we can see that our method is better. 

\begin{table}[tbp]
\begin{center}
  \caption{ Parameters and FLOPs comparison}
  \label{tab:flop_param}
  \begin{tabular}{l|l|l|l|l|l}
   \hline
   \hline
    & RCAN   & MSRN & DBPN   & HRAN   & Ours  \\ 
    \cline{2-6}
   Param  & 22.89M & 5.67M & 10.04M & 8.01M & \textbf{2.24M} \\ 
   \cline{2-6}
   FLOPs  & 29.96G (52.30G)  & 97.3G & 1438.26G & 170G & \textbf{4.95G} \\ \hline \hline
   
  \end{tabular}
  
\end{center}
 \end{table}

\begin{table}[tbp]
\centering
\caption{The results of ablation studies on B100(x4). The ‘ $\times$ ’ in this table denotes the corresponding operation is not used and the ‘ $\surd$ ’ denotes the corresponding operation is used.}
\label{tab:ablation}
\begin{center}
\resizebox{0.7\linewidth}{!}{
\begin{tabular}{c|c|c|c|c|c}
	\toprule
	\hline
                  & ab\_1 & ab\_2 & ab\_3  & ab\_4  & ab\_5     \\
    \cline{2-6}
Standard Conv     & $\surd$       & $\times$   & $\times$      & $\times$    & $\times$   \\
Ghost Module      & $\times$      & $\surd$    & $\surd$       & $\surd$     & $\surd$   \\
Channel Attention & $\surd$       & $\surd$    & $\surd$      & $\times$    & $\times$   \\
Spatial Attention & $\times$      & $\times$   & $\surd$       & $\surd$    & $\times$  \\
Param             & 22.89M      & 2.21M    & 2.24M    & 2.07M    & 2.04M   \\
FLOPs             & 52.30G      & 4.88G    & 4.95G    & 4.95G   & 4.88G   \\
avg\_PSNR         & 27.77       & 27.54    & 27.84    & 27.47   & 27.06   \\
avg\_SSIM         & 0.7436      & 0.7146   & 0.7276   & 0.7214 & 0.7114    \\
\hline
\bottomrule
\end{tabular}}
\end{center}
\end{table}

\begin{figure}
	\centering
	\includegraphics[width=0.95\textwidth]{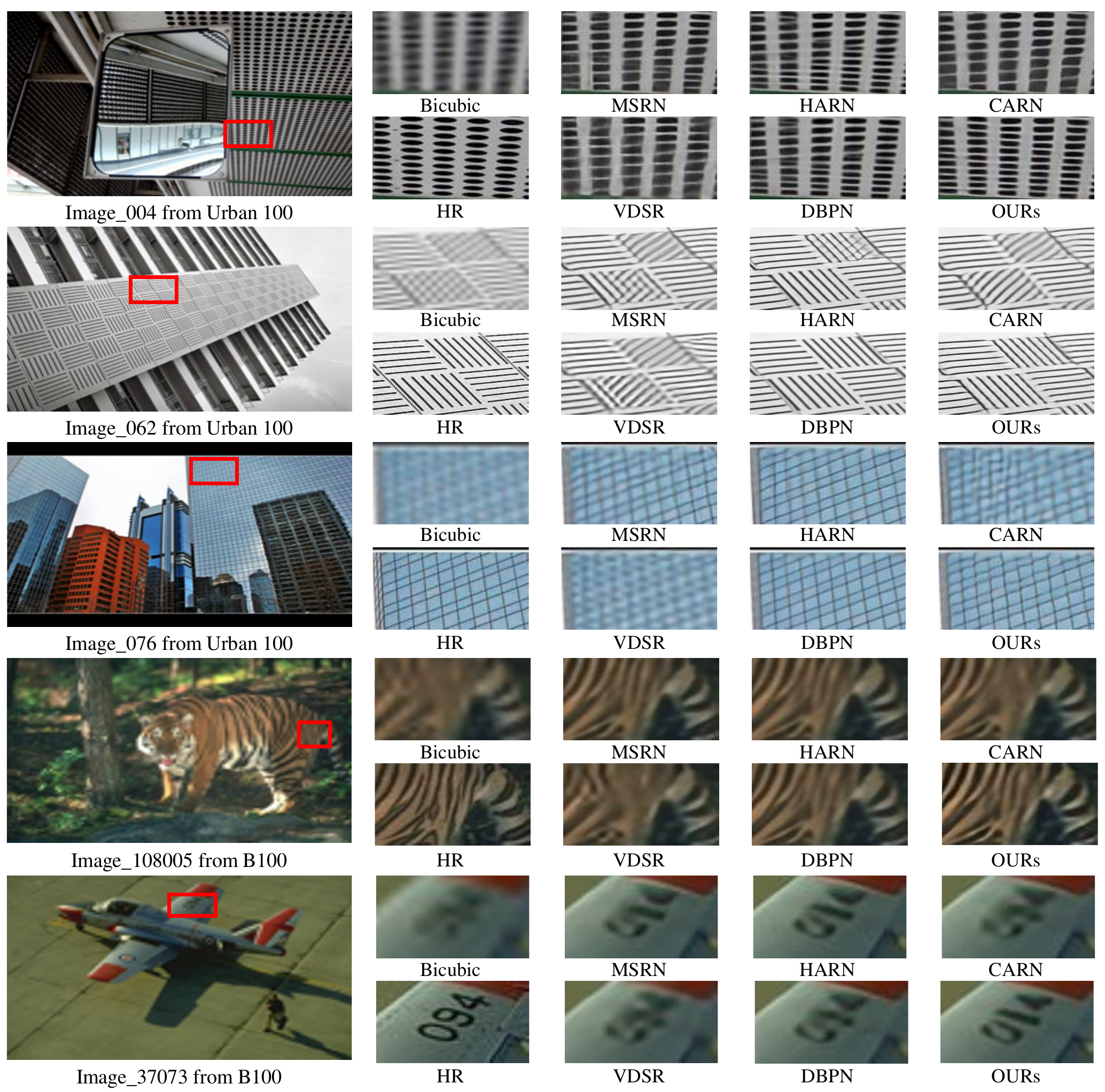}
	\caption{ Visual comparison with bicubic degradation model.}
	\label{results}
\end{figure}

{\bfseries Qualitative Visual comparison.}
We also show visual comparison results in Figure~\ref{results}. The leftmost column is the input LR of the algorithm, and the four columns on the right, except the HR (Ground Truth), are the output results, SR, of the partial comparison algorithm, corresponding to the magnification of the red box in LR. 
The results show that our method has a certain degree of advancement in image information reconstruction. This improvement can be clearly seen in the enlarged color patches, especially in terms of details.
Such as the ‘Image004’ and ‘Image076’ from Urban 100, which show our method has an obvious effect on line restoration. 

\begin{figure}
	\centering
    \includegraphics[width=0.95\textwidth]{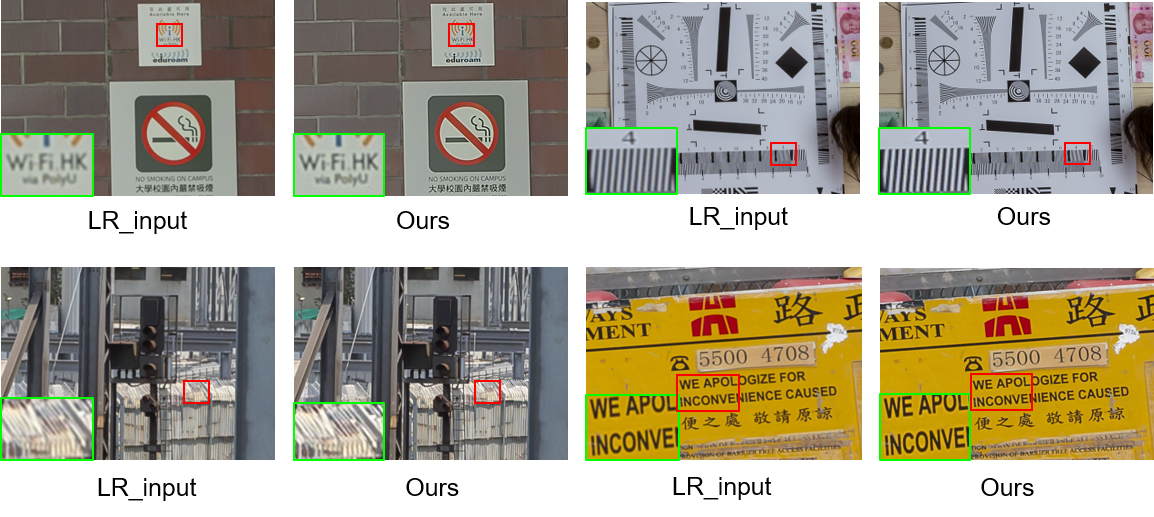}
	\caption{ Experimental results on real data.}
	\label{fig8}
\end{figure}

{\bfseries Evaluation on real data.} We further conduct experiments on real-world LR images to demonstrate the effectiveness of our method. We adopt the new RealSR~\cite{cai2019toward} dataset, which is used in the NTIRE2019 competition~ \cite{cai2019ntire}. This dataset contains raw images captured by DSLR cameras. Multiple images of the same scene have been captured with different focal lengths. Images taken with longer focal lengths contain finer details and can be considered as HR counterparts for images with shorter focal lengths. Although RealSR provides images on different scales, it is really hard to obtain image pairs that are totally aligned. That is because of complicated misalignment between images and changes in the imaging system introduced by adjusting the focal length. As a result, we only consider a visual comparison of SR results. We use images captured with 28mm focal length as LR inputs, and Figure \ref{fig8} shows the visual comparison of 2x upsampling on the RealSR dataset, by which we can find that our method has achieved good visual effects.  

\subsection{Ablation Study}
To analyze the contribution of the proposed GRAB and CSAM for our lightweight model GRAN, we compare our method with the ablations of various versions. Actually, our GRAN consists of multiple GRAB groups, and each GRAB consists of Ghost Module and Channel and Spatial Attention Mechanism. Therefore, our ablation study expands by validating each part in GRAB, \ie Ghost Module, channel attention, and spatial attention. Here we name the original convolution operation in the traditional network mentioned in Section~\ref{sec:ghost_module} as standard convolution. As shown in Table~\ref{tab:ablation}, we designed 5 versions of the experimental setup for ablation:\\
\textbf{ab1} First, we adopt standard convolution operation and channel attention mechanism to build up the basic block. Actually, it equals the basic block in RCAN~\cite{zhang2018image},~\ie RCAB. \\
\textbf{ab2} For verifying the effect of standard convolution and Ghost Module on the number of parameters, the amount of computation, and the model's performance. We adopt Ghost Module, a channel attention mechanism, to build up the basic block. The results in Table~\ref{tab:ablation} reveal that the Ghost Module significantly reduces the number of parameters and calculations of the model compared with RCAN~\cite{zhang2018image}. \\
\textbf{ab3} We adopt Ghost Module, Channel, and Spatial Attention Mechanism (CSAM) to build up the basic block,~\ie GRAB used for constructing our GRAN. Compared with~\textbf{ab2},~\textbf{ab3} further proves that our designed CSAM contributes to improved performance.\\
\textbf{ab4} In addition, to verify the influence of the Channel and Spatial Attention Mechanism on the number of parameters, the amount of computation, and the performance of the model. We remove the Channel Attention.\\
\textbf{ab5} Finally, we continue to move off the spatial Attention on the basis of \textbf{ab4}. Compared with~\textbf{ab3}, when only channel attention or only spatial attention is used or both not though the parameters of the models are approximately equal, both PSNR and SSIM have significantly decreased. 

The above ablation studies indeed reflect the effectiveness of the proposed GRAB and CSAM. That is to say, the Ghost Module using linear operation can indeed reduce the complexity of the network compared with traditional convolution, and the attention mechanism that mixes space and channel can also help to rebuild a higher-quality image.

\section{Conclusion}
In this paper, we propose an efficient and lightweight network named GRAN for SISR. It consists of multiple GRAB groups with Ghost Technology and Channel and Spatial Attention Mechanism to reduce redundant features and decrease memory and computing resource requirements. Experiments conducted on the benchmarks show the superior performance of our method in qualitative and quantitative. We achieve higher performance with lower computational resources, whose parameters have decreased by more than ten times compared to the previous SoTA methods. Currently, we can handle low-resolution images well. However, the proposed method may fail when dealing with more realistic scenarios, such as blur and noise in low-resolution images. Therefore, we will focus on solving the super-resolution problem under complex degradation in the future.

\section*{Declarations}

\begin{itemize}
\item Funding

This work was funded in part by the Project of the National Natural Science Foundation of China under Grant 61901384 and 61871328, the Natural Science Basic Research Program of Shaanxi under Grant 2021JCW-03, as well as the Joint Funds of the National Natural Science Foundation of China under Grant U19B2037.
\item Conflict of interest/Competing interests (check journal-specific guidelines for which heading to use)

Not applicable

\item Ethics approval 
\item Consent to participate
\item Consent for publication
\item Availability of data and materials

The datasets generated during and/or analyzed during the current study are available from the corresponding author on reasonable request.

\item Code availability 

After the paper is accepted, the code will be open-sourced.
\item Authors' contributions

Conceptualization, A.N, and Q.Y.; methodology, A.N.; software, A.N,  P.W.; validation, A.N., P.W.; formal analysis, A.N.; investigation, P.W.; resources, Y.Z.; data curation, J.S.; writing---original draft preparation, A.N.; writing---review and editing, A.N., P.W., Y.Z, S.J.; visualization, P.W.; supervision, S.J.; project administration, A.N., Y.Z. All authors have read and agreed to the published version of the manuscript.
\end{itemize}

\bibliography{sn-bibliography}

\end{document}